\documentclass[letterpaper,10pt,conference]{ieeeconf}
\IEEEoverridecommandlockouts

\usepackage{amssymb}
\usepackage{booktabs}
\usepackage{color}
\usepackage[backend=biber,
            url=false,
            isbn=false,
            doi=false,
            backref=false,
            style=ieee,
            natbib=true,%compatibility aliases
            mincitenames=1,
            maxcitenames=1,
            citestyle=numeric-comp,
            sorting=none,%none
            block=none]{biblatex}
\renewcommand{\bibfont}{\small}
\usepackage{graphicx}
\let\labelindent\relax
\usepackage{enumitem}
\usepackage{amsmath}
\usepackage{wrapfig}
\usepackage{multirow}
\usepackage{caption}
\usepackage{subcaption}
\usepackage{tabularx,array}
\usepackage{listings}
\usepackage{xcolor}
\usepackage{tcolorbox}
\usepackage{xspace}

\usepackage[colorlinks=true,
            linkcolor=blue,
            citecolor=blue,
            urlcolor=blue]{hyperref}
\usepackage[nameinlink, capitalize]{cleveref} % for cref figure

\usepackage{siunitx}
\usepackage{caption}
\newcommand{\ACRO}{{MIDAS Hand}\xspace}

\definecolor{lightgray}{gray}{0.95}

\lstdefinelanguage{yaml}{
  morekeywords={true,false,null,yes,no},
  sensitive=false,
  morecomment=[l]{\#},
  morestring=[b]",
  morestring=[b]'
}

\begin{document}
% \overrideIEEEmargins
    
\title{\ACRO: Modular low-Impedance Directly-driven Anthropomorphic Sensing Hand}

\author{%
Alvin Zhu$^{1, 2*}$, Mingzhang Zhu$^{3*}$, Quanyou Wang$^{3*}$, Beom Jun Kim$^{3*}$, Jose Victor S. H. Ramos$^{3}$, Dennis Hong$^{3}$ \\
\href{https://midas-hand.com}{midas-hand.com}
\thanks{* denotes equal contribution.}
\thanks{$^{1}$Department of Computer Science, $^{2}$Department of Electrical and Computer Engineering, $^{3}$Department of Mechanical and Aerospace Engineering, UCLA, Los Angeles, CA, USA.}
}

\maketitle

\begin{abstract}
Dexterous manipulation is limited not only by algorithms but by a shortage of accessible hand hardware that combines human-scale morphology, ease of manufacturing or maintenance, tactile sensing, and practical cost. Existing dexterous hands tend to optimize some of these properties at the expense of others. We present \ACRO, a low-cost, open-source, human-scale dexterous hand with integrated tactile sensing for manipulation research. \ACRO provides 16 total degrees of freedom (DoF) with 13 active DoF, directly-driven actuation with measurably low backdrive torque, and 283 three-axis tactile taxels in a compact 700 g package with a bill of materials under 3,000 USD. Built from 3D-printed components, it assembles in under three hours while providing the strength, repeatability, and maintainability needed for repeated real-world experiments. Alongside the hardware, we release a full stack: design files, build documentation, control and tactile python APIs, simulation models, and retargeting and teleoperation pipelines. We characterize \ACRO through workspace and grasp-taxonomy analysis, payload and reliability tests, backdrivability measurements, and teleoperation demonstrations with tactile sensing, showing that it offers a balanced, reproducible platform for tactile dexterous manipulation and human-to-robot data collection. More details are available at \url{https://midas-hand.com}.

\end{abstract}

\section{Introduction}

Dexterous manipulation is challenging since many everyday tasks require more than stable grasping: a hand must adjust contacts, regulate interaction forces, manipulate objects under uncertainty, and operate tools built for the human hand. Recent dexterous learning, teleoperation, and demonstration-collection methods have produced increasingly capable behaviors, but they are data-hungry and remain strongly constrained by the hand hardware used to collect data and execute policies~\cite{xu2025dexumi,zhu2026dexexowearabilityfirstdexterousexoskeleton,zheng2026egoscalescalingdexterousmanipulation,tang2026humanlikemanipulationrlaugmentedteleoperation}. The hand is thus not only an end-effector, but also a sensing platform, a data-collection device, and the physical interface between algorithms and the real world.

\begin{figure}[htbp!]
 \centering
 \includegraphics[width=\linewidth]{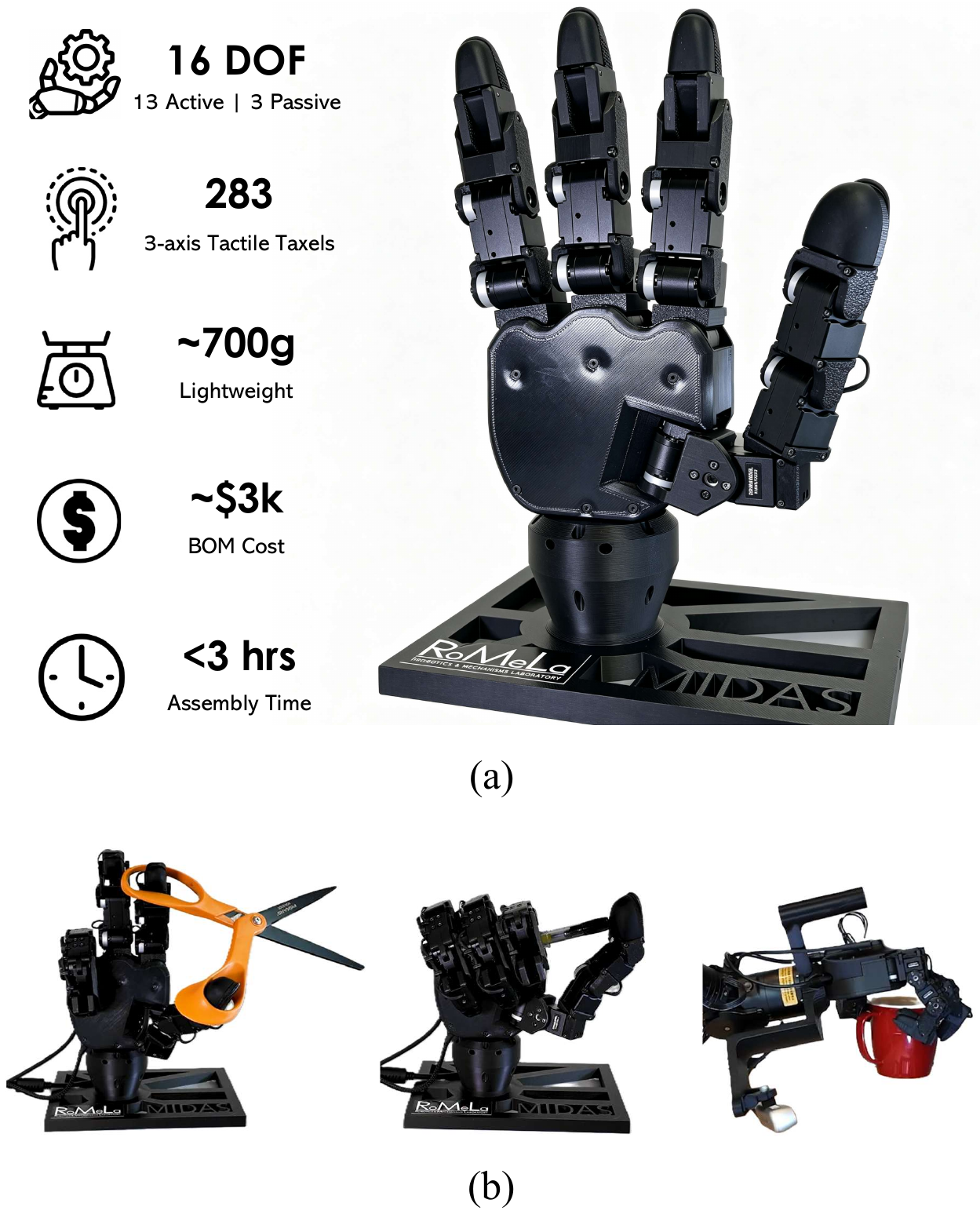}
 \captionsetup{belowskip=-10pt}
 \caption{(a) \ACRO and its key specifications. (b) Representative grasps: scissor operation, pen use, and irregular-object grasping.}
 \label{fig:main}
\end{figure}

These roles impose competing requirements: a research hand must be dexterous enough for contact-rich tasks, instrumented to sense its contacts, robust for repeated trials, affordable, and reproducible. Human-scale morphology and tactile sensing matter most. Because objects and tools are built around the size, workspace, and opposition of the human hand, comparable morphology simplifies interaction, teleoperation, retargeting, and human-to-robot skill transfer, and thumb opposition and finger workspace are tied to precision and fine manipulation~\cite{KUO2009829,sym14010071,kang2026kinematicoptimizationphalanxlength,peticco2026karmakinematicmetricfine}. Tactile feedback supplies contact, force, shear, and interaction-state information that vision cannot observe under occlusion. By enabling slip detection, grasp-stability monitoring, and contact-state estimation, it is increasingly central to tactile-reactive control and visuotactile learning~\cite{niu2026trextactilereactivedexterousmanipulation,heng2025vitacformerlearningcrossmodalrepresentation,cheng2025omnivtla,vtam2026,song2025opentouchbringingfullhandtouch}.

No existing hand jointly satisfies these requirements. Commercial hands offer mature integration and high dexterity but remain costly, closed-sourced, and hard to modify~\cite{Allegro2024,wujihand-2025,sharpawave2025}, while open-source hands improve accessibility yet individually trade away human-scale morphology, dense tactile sensing, robustness, or maintainability~\cite{shaw2023leaphand,zorin2025ruka,liu2026rukav2tendondrivenopensource,christoph2025orca,lin2026crafttendondrivenhandhybrid}. Dense tactile sensing in particular is difficult to integrate into a compact, low-cost, reproducible hand, precisely where contact-rich manipulation needs it most. Accessible hardware is also only part of the solution: without control and tactile APIs, simulation models, and retargeting and teleoperation tools, a hand still demands substantial integration effort before it can support demonstration collection and learning.

We present \ACRO, a low-cost, open-source, human-scale dexterous hand with integrated tactile sensing for both real-world experiments and scalable demonstration collection. It provides 16 total DoFs with 13 active DoFs across four fingers, an opposable thumb, and 283 three-axis tactile taxels, all in a 700~g package under USD~3,000. Its actuation is directly-driven, with each actuator driving its joint through rigid linkages rather than tendons, and its low-friction transmission yields measurably low backdrive torque for compliant interaction. The modular 3D-printed structure assembles in under three hours with easy maintenance. We release the full stack including design files, control and tactile APIs, simulation models, retargeting and teleoperation pipelines, and characterize the hand through kinematic, backdrivability, tactile, reliability, and manipulation experiments.

\textbf{Contributions.} The contributions of this work are:
(i) a compact, human-scale dexterous hand that reaches a balance prior open hands achieve only individually: directly-driven backdrivable actuation, dense multi-axis tactile sensing, and low-cost 3D-printed construction.
(ii) a full-stack, open-source software and data-collection ecosystem (control and tactile APIs, auto homing, simulation models, retargeting, and teleoperation) usable beyond the originating lab.
(iii) experimental validation across workspace and grasp-taxonomy coverage, payload, reliability, backdrivability, and tactile and teleoperation demonstrations.
\section{Related Work}
Commercial hands such as the Allegro Hand, Wuji Hand, and Sharpa Wave provide mature hardware, but their high cost, proprietary designs, and restricted access to hardware and software internals make them difficult to modify, deploy in academic labs, or furthur develop~\cite{Allegro2024,wujihand-2025,sharpawave2025}. Among open-source platforms, LEAP Hand is low-cost and anthropomorphic but larger than a human hand and lacks tactile sensing~\cite{shaw2023leaphand}, while RUKA, RUKA-v2, and ORCA are tendon-driven and lack high-resolution tactile feedback~\cite{zorin2025ruka,liu2026rukav2tendondrivenopensource,christoph2025orca}; CRAFT adds hybrid hard-soft compliance but is likewise tendon-driven~\cite{lin2026crafttendondrivenhandhybrid}. Directly-driven actuation offers several advantages over tendon-driven designs: it is structurally simpler and easier to maintain, transfers more readily from simulation to hardware, and keeps all actuators inside the hand rather than a separate forearm module, allowing the hand to attach directly from the wrist without adding unnecessary forearm length to commercial 6- and 7-DoF arms. These actuation benefits are shared by other directly-driven hands such as Wuji Hand and Sharpa Wave. \ACRO pairs directly-driven actuation with human-scale morphology, low backdrive torque, integrated high-resolution tactile sensing, low cost, and modular maintainability in a single reproducible platform.

\subsection{Tactile sensing for dexterous manipulation}
Tactile sensing is increasingly essential for robust contact-rich manipulation research. Tactile-reactive and visuotactile learning systems show that contact feedback provides information that is hard to infer from vision~\cite{niu2026trextactilereactivedexterousmanipulation,heng2025vitacformerlearningcrossmodalrepresentation,cheng2025omnivtla,vtam2026}, and studies on full-hand systems demonstrate that broader tactile coverage enhances real-world interaction and physical reasoning~\cite{song2025opentouchbringingfullhandtouch}. Yet dense tactile integration in a compact hand poses mechanical, electrical, and maintenance challenges, and hands that achieve it are often costly or hard to reproduce. \ACRO does not propose a new sensor; it integrates commercial multi-axis tactile arrays~\cite{paxini2026} into a low-cost, reproducible hand as a default modality, validated through tactile sensing experiments. 

\subsection{Teleoperation, retargeting, and data collection}
Dexterous manipulation also depends on software: simulation, hardware APIs, retargeting, and teleoperation pipelines lower the effort to collect demonstrations and transfer policies. DexPilot, AnyTeleop, and contact-rich retargeting map human motion onto robotic hands~\cite{handa2020dexpilot_arxiv,qin2023anyteleop,lakshmipathy2024kinematicmotionretargetingcontactrich}, while recent work scales teleoperation and data collection through haptic gloves, egocentric data, EMG-based pose estimation, and high-DoF retargeting~\cite{ZhangH-RSS-25,zheng2026egoscalescalingdexterousmanipulation,zhao2026dexemgdexterousteleoperationemg2pose,bytedance2025bytedexter,rayyan2026teledexaccessibledexterousteleoperation}. These efforts show that data-driven manipulation needs not only retargeting but sensorized, reproducible hands with accessible APIs and synchronized streams, which \ACRO provides by releasing hardware, simulation, retargeting, and teleoperation tools together.
\section{System Design}
\label{sec:system}

\begin{table*}[htbp!]
\centering
\captionsetup{justification=centering}
\caption{Comparison of open-source and commercially available dexterous robotic hands.}
\label{tab:compare}
\renewcommand{\arraystretch}{1.35}
\setlength{\tabcolsep}{4pt}
\scriptsize
    \begin{tabular*}{\textwidth}{@{\extracolsep{\fill}}lccccccc@{}}
    \toprule
    \multirow[c]{2}{*}[-3em]{\textbf{Hands}} &
    \textbf{LEAP~\cite{shaw2023leaphand}} &
    \textbf{RUKA v2~\cite{liu2026rukav2tendondrivenopensource}} &
    \textbf{ORCA~\cite{christoph2025orca}} &
    \textbf{Allegro v5 Plus~\cite{Allegro2024}} &
    \textbf{Sharpa Wave~\cite{sharpawave2025}} &
    \textbf{Wuji Hand~\cite{wujihand-2025}} &
    \textbf{\ACRO (Ours)} \\
 
    &
    \includegraphics[width=0.8cm]{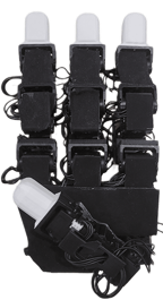} &
    \includegraphics[width=1.6cm]{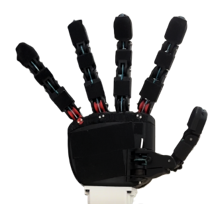} &
    \includegraphics[width=1.8cm]{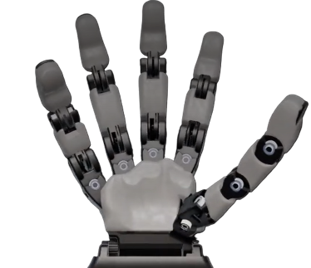} &
    \includegraphics[width=1.7cm]{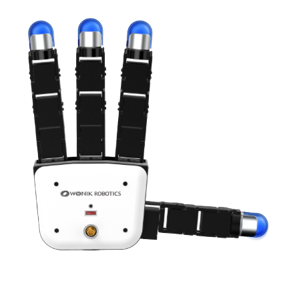} &
    \includegraphics[width=1.5cm]{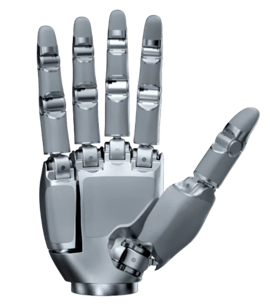} &
    \includegraphics[width=1.5cm]{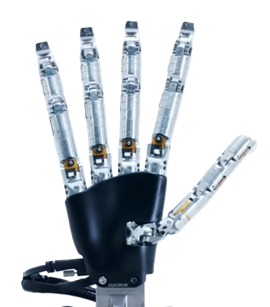} &
    \includegraphics[width=1.5cm]{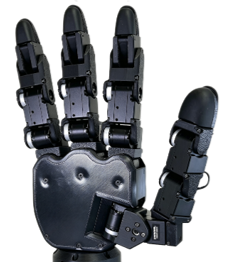} \\
    \hline
    
    \textbf{Total DoF} & 
    16 & 
    20 & 
    17 & 
    16 & 
    22 & 
    20 & 
    16 \\  
    
    \textbf{Actuation} &
    Direct &
    Tendon &
    Tendon &
    Direct &
    Direct &
    Direct &
    Direct \\
    
    \textbf{Tactile Sensor} &
    No &
    No &
    Yes (BTF) &
    Yes (Pneumatic) &
    Yes (DTA) &
    No &
    Yes (DTA) \\
    
    \textbf{Size} &
    $\sim$30\% larger &
    Human-scale &
    Human-scale &
    $\sim$30\% larger &
    Human-scale &
    Human-scale &
    Human-scale \\
    
    \textbf{Weight (g)} &
    $\sim$1,000 &
    N.A. &
    $\sim$1,200 &
    1,240 &
    $\sim$1,300 &
    $<$600 &
    $\sim$700 \\
 
    \textbf{Cost (USD)} &
    $\sim$2,000 &
    $\sim$1,500 &
    $\sim$2,500 &
    $\sim$11,000 &
    $\sim$50,000 &
    $\sim$16,000 &
    $<$3,000\\
    \bottomrule
    \end{tabular*}
    \par\vspace{2pt}
    {\footnotesize BTF: binary tactile feedback. DTA: dynamic tactile array. % NOTE: confirm "binary" is intended for MIDAS/Sharpa — the Paxini taxels report continuous 3-axis force per taxel, so "binary" may contradict the 283 three-axis taxel claim.
    }
\end{table*}

\ACRO is designed as a low-cost, human-scale dexterous hand platform for manipulation research. As summarized in Table~\ref{tab:compare}, the design aims to balance several practical requirements that are often difficult to achieve simultaneously, including anthropomorphic size, dexterity, integrated tactile sensing, low weight and cost, and ease of fabrication and maintainability. The hand features a four-finger configuration with 16 total DoFs and 13 active DoFs, including a four-DoF thumb and three opposing fingers. The mechanical structure is primarily fabricated from 3D-printed components, allowing the hand to remain lightweight and low-cost while supporting repeated real-world manipulation experiments. In addition, \ACRO integrates tactile sensing and provides supporting electronics and software infrastructure for calibration, simulation, retargeting, and teleoperation. The following sections describe the mechanical design, electronics, and software stack in detail.

\subsection{Actuation and transmission}

% finger config
\begin{figure}[htbp!]
    \centering
    \includegraphics[width=0.95\linewidth]{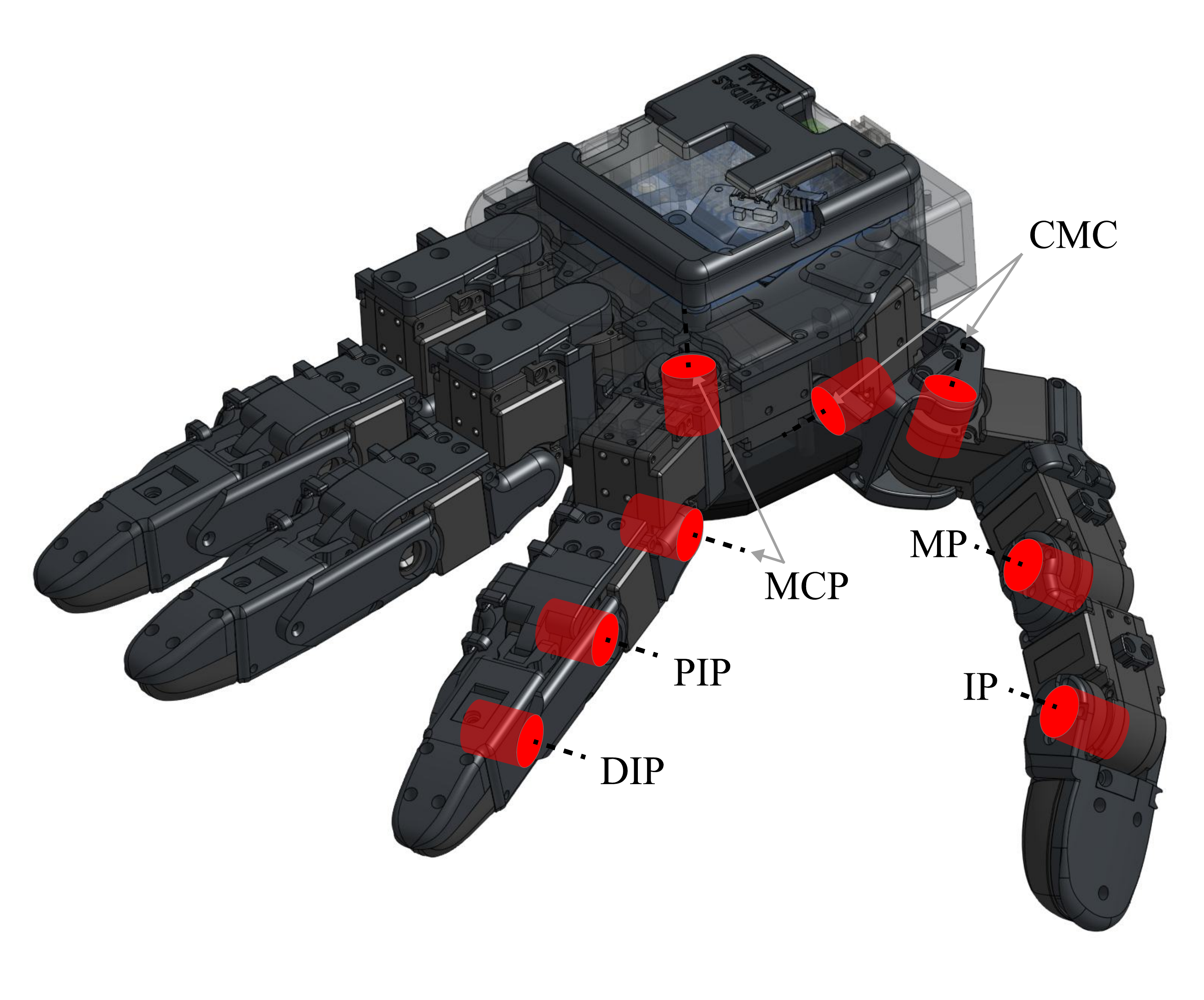}
    \captionsetup{belowskip=-10pt}
    \caption{Kinematic structure and joint nomenclature of the \ACRO.}
    \label{fig:joint_config}
\end{figure}

\ACRO drives all 13 active joints with identical high-performance Dynamixel XM335-T323-T servomotors. The actuation system is embedded within a compact structure, leveraging directly-driven actuation with rigid linkages for transmission. This approach provides high torque density and precise position control, while the low-friction transmission keeps the joints backdrivable for compliant, robust interaction with diverse objects. The resulting backdrive torque is characterized in Sec. IV-A. 

\subsection{Finger Configuration \& Kinematics}
To achieve anthropomorphic grasping capabilities, the finger configurations of \ACRO are designed based on human anatomical and clinical data \cite{li2019anthropometric, karanjkar2024normal}, with the joint nomenclature defined in \cref{fig:joint_config}. The overall hand dimensions (205×120×55 mm), measured on the physical \ACRO prototype, closely match the average adult human hand, ensuring seamless integration into human-centric environments and teleoperation setups.

\begin{figure}[htbp!]
    \centering
    \includegraphics[width=0.95\linewidth]{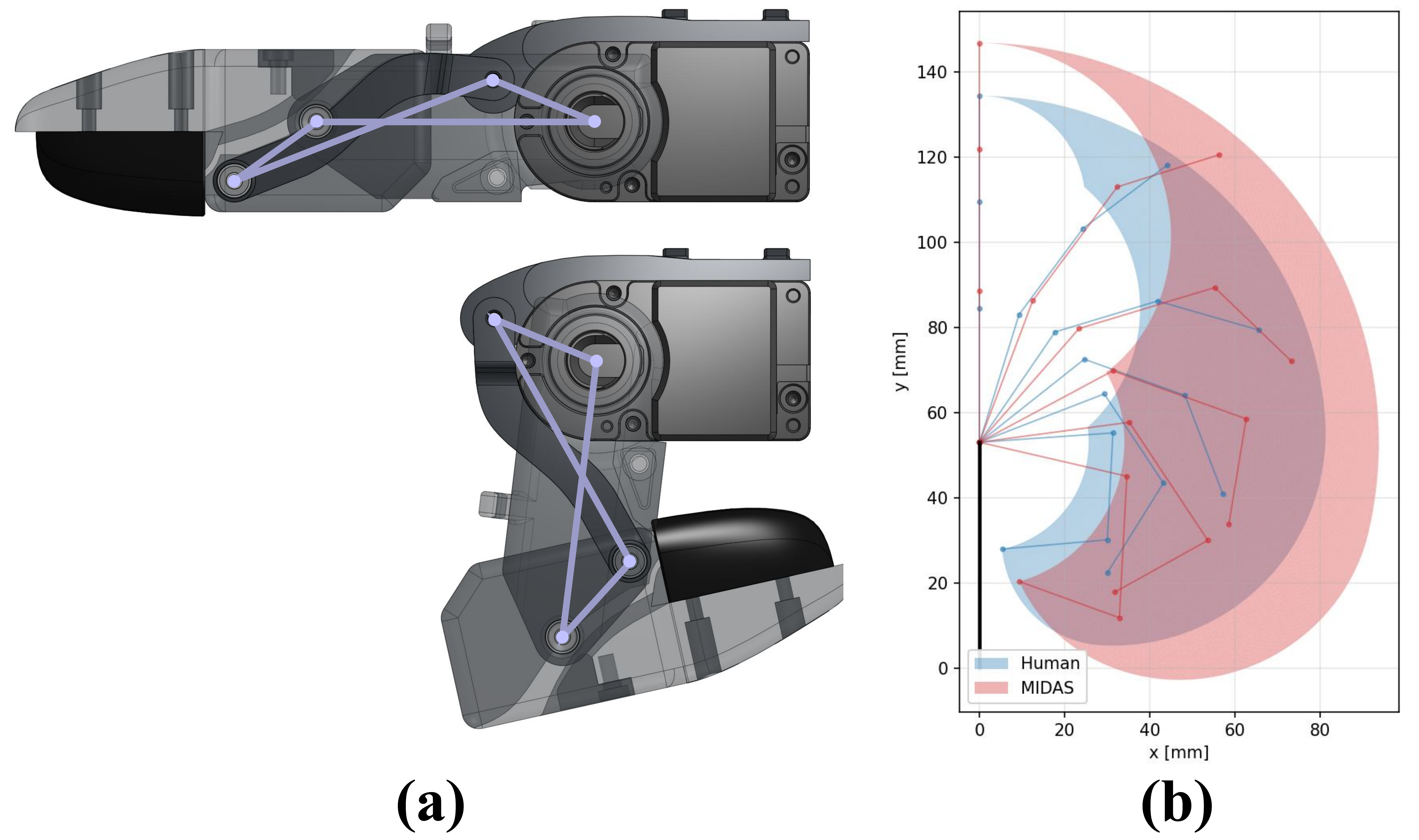}
    \captionsetup{belowskip=-10pt}
    \caption{Fingertip ROM and linkage mechanism of \ACRO. (a) Side view of the passive DIP linkage coupling PIP and DIP motion. (b) Sagittal-plane workspace of the \ACRO finger (red) versus a human index finger (blue).}
    \label{fig:rom}
\end{figure}

The index, middle, and ring fingers utilize an underactuated architecture driven by a crossed four-bar linkage mechanism for the Distal Interphalangeal (DIP) joint. As illustrated in \cref{fig:rom} (a), this kinematic coupling enforces a deterministic relationship between the DIP and Proximal Interphalangeal (PIP) joints, allowing simultaneous flexion through a single actuator. This approach significantly reduces control complexity while improving compliance and structural reliability during power grasps. The resulting sagittal-plane workspace of the index finger is evaluated in \cref{fig:rom} (b). Assuming equal overall hand length between the robot and a human hand, and scaling human index-finger proportions from population anthropometric data \cite{li2019anthropometric}, the \ACRO finger covers 80.1\% of the human index finger's sagittal-plane workspace, computed as the fraction of the human workspace area also reachable by the robot finger; the excess robot workspace reflects its proportionally longer index finger relative to the human population average.

The thumb features 4 active DoF to replicate a human-like opposition workspace. Instead of mounted orthogonally to the palm, the thumb incorporates an oblique second joint axis. This design ensures that its neutral orientation closely aligns with which of human biological counterpart. This kinematic configuration enables the thumb to sweep across the palm and directly oppose the other fingertips, supporting a diverse repertoire of grasping modes (pinch, power grasps, and in-hand object stabilization) within a compact, human-scale form factor. The full joint range of motion (ROM) is summarized in Table~\ref{tab:midas_rom}.

\subsection{Modularity and Assembly}
The index, middle, and ring fingers are identical, self-contained modules, each integrating its three actuators, the crossed four-bar DIP linkage, and a fingertip tactile module. Each module mounts to the palm through a few M2 screws and two wire connectors, so removing a finger requires no internal rewiring. As a result, the complete hand can be assembled from its 3D-printed components in under three hours, and a finger module can be swapped in under 15 minutes on average. This modularity simplifies maintenance and repair, supporting sustained use across repeated real-world experiments.

\subsection{Tactile sensing}
To support contact-rich manipulation, \ACRO integrates high-density commercial tactile sensing arrays on the thumb and fingertips. The index, middle, and ring fingertips use Paxini PX6AX-GEN3-DP-S2015-Elite tactile modules, while the thumb uses the larger PX6AX-GEN3-DP-M2826-Omega module to provide a larger sensing area and higher load capacity for thumb-opposition grasps. Each S2015-Elite module provides 52 three-axis taxels, while the M2826-Omega provides 127 taxels, for a total of $3 \times 52 + 127 = 283$ taxels across the hand. The sensors are mounted on the primary contact surfaces of the thumb, index, middle, and ring fingers, allowing the hand to measure local contact location and force distribution during grasping and manipulation.
 
\begin{figure}[t]
    \centering
    \includegraphics[width=0.95\linewidth]{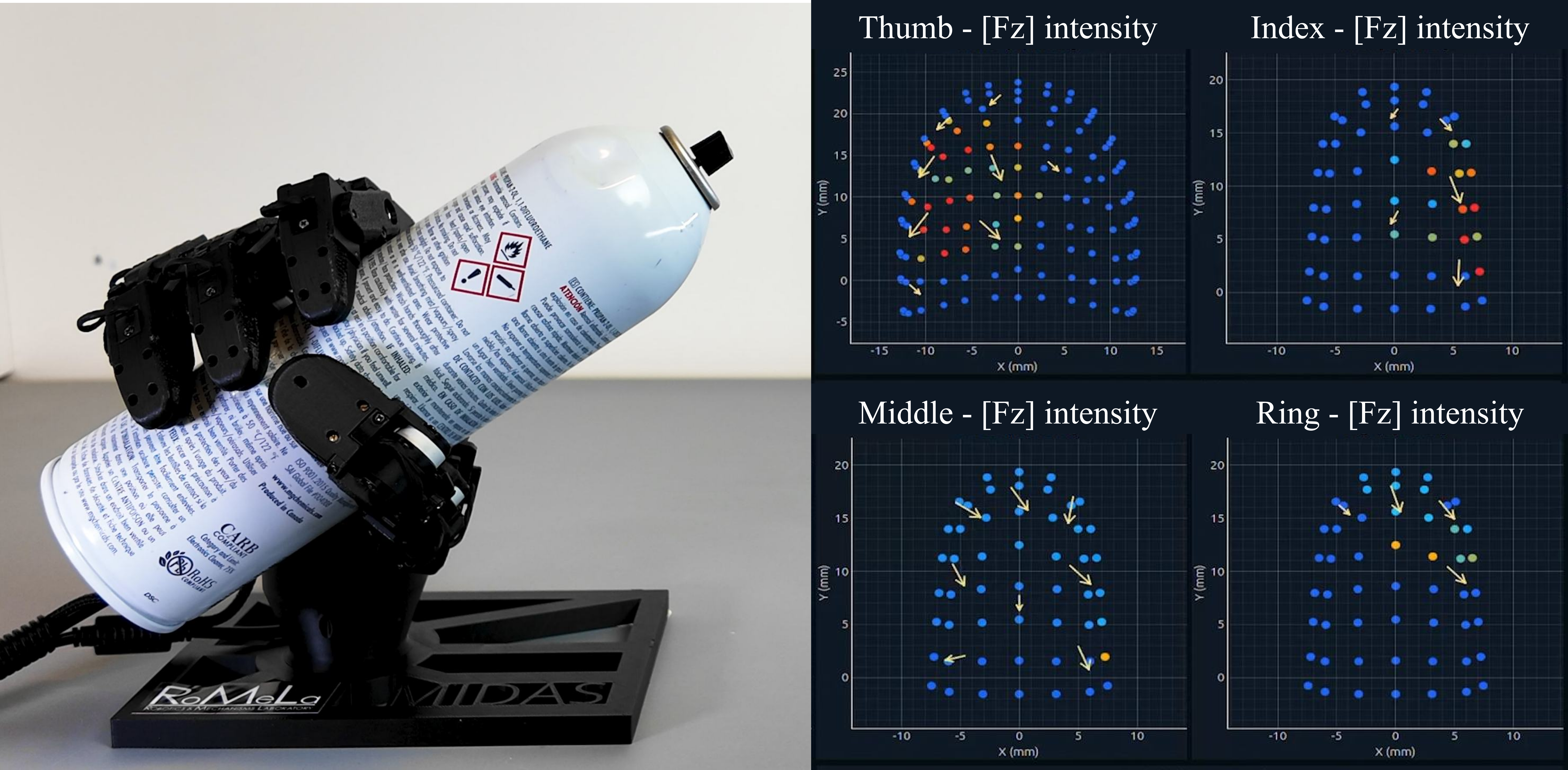}
    \captionsetup{belowskip=-10pt}
    \caption{Tactile sensor readings during object grasping.}
    \label{fig:tactile}
\end{figure}

According to the manufacturer-reported specifications, each tactile module provides multi-axis force sensing with normal and shear force measurements. The sensors support a normal force range up to $25\,\mathrm{N}$ and a tangential force range up to $\pm 10\,\mathrm{N}$, with a maximum output rate of $83.3\,\mathrm{Hz}$. The reported minimum recognition force is $0.1\,\mathrm{N}$, and the spatial resolution is $1\,\mathrm{mm}$. These characteristics allow \ACRO to capture contact position, normal loading, and shear information as shown in \cref{fig:tactile}, which are important for stable grasping, contact monitoring, and future tactile feedback control. The tactile sensors are connected through the vendor-provided high-speed communication board and streamed to the host computer together with actuator state feedback.
 
% tactile

\subsection{Electronics}
The electronic architecture of \ACRO is designed to provide a compact and maintainable interface for actuator control and tactile data acquisition. A custom distribution board located in the palm routes power and TTL communication to the Dynamixel actuators. Each finger module connects directly to this board, thereby enabling all actuators to share a common TTL communication bus. This layout reduces wiring complexity, simplifies assembly and debugging, and allows individual finger modules to be replaced or serviced independently. A U2D2 power hub provides the external power interface and communication link between the hand and the host computer. Tactile sensing is handled through Paxini’s high-frequency communication board, which interfaces with the tactile sensors and streams contact measurements to the host system. By separating actuator distribution, tactile acquisition, and host communication into dedicated modules, the electronics remain compact, modular, and easy to maintain for repeated manipulation experiments.

\begin{table}[t]
\centering
\caption{\ACRO Joint ROM}
\label{tab:midas_rom}
\begin{tabular}{llc}
\toprule
\textbf{Finger} & \textbf{Joint Name} & \textbf{Limits [deg]} \\
\midrule
Thumb & CMC Roll & $[0, 123]$ \\
      & CMC Side & $[-45, 52]$ \\
      & MP      & $[-90, 90]$ \\
      & IP       & $[-90, 90]$ \\
\midrule
Index, Middle, Ring & MCP Ab/Ad & $[-45, 45]$ \\
      & MCP Pitch & $[-103, 0]$ \\
      & PIP       & $[-83, 0]$ \\
      & DIP*      & $[-107, 0]$ \\
\bottomrule
\end{tabular}
\par\vspace{2pt}
{\footnotesize $^{*}$Passive joint, coupled to the PIP through the four-bar linkage.}
\end{table}

\subsection{Open-Source Ecosystem}
\begin{figure}[htbp!]
    \centering
    \includegraphics[width=0.99\linewidth]{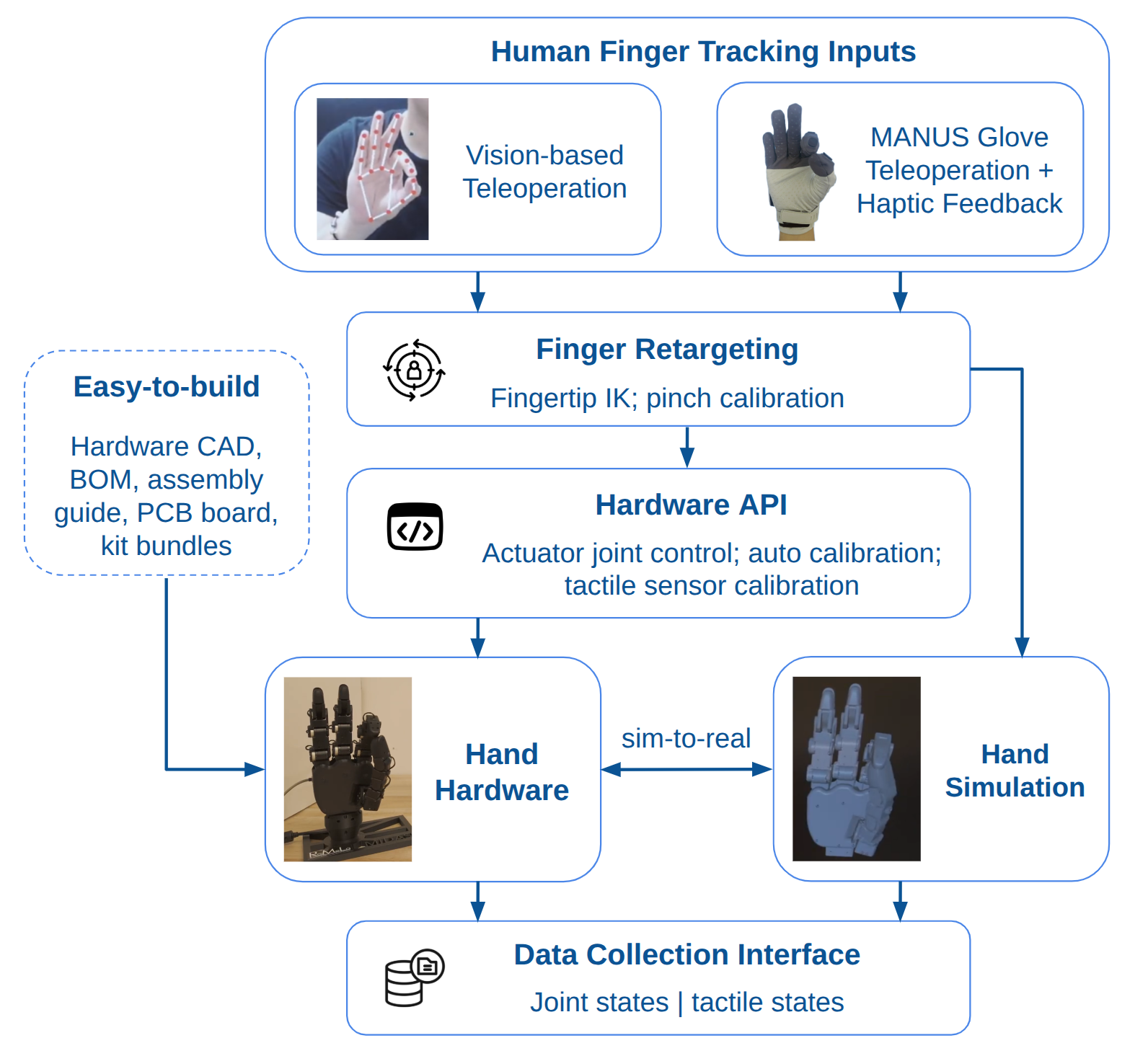}
    \captionsetup{belowskip=-5pt}
    \caption{Overview of the \ACRO open-source ecosystem: hardware release artifacts, teleoperation inputs, retargeting, hardware API, simulation, and data collection.}
    \label{fig:ecosystem}
\end{figure}

To make \ACRO reproducible beyond the originating lab, we release the complete ecosystem summarized in \cref{fig:ecosystem}. On the hardware side, the release includes the full CAD models on Onshape, downloadable STEP files, bill of materials, PCB design files, and a step-by-step assembly guide, together with tactile sensor and actuator kit bundles that simplify part sourcing. This combined with the modular 3D-printed structure allow the hand to be built and serviced without specialized tooling or complex custom components.
 
\begin{figure}[htbp!]
    \centering
    \includegraphics[width=0.9\linewidth]{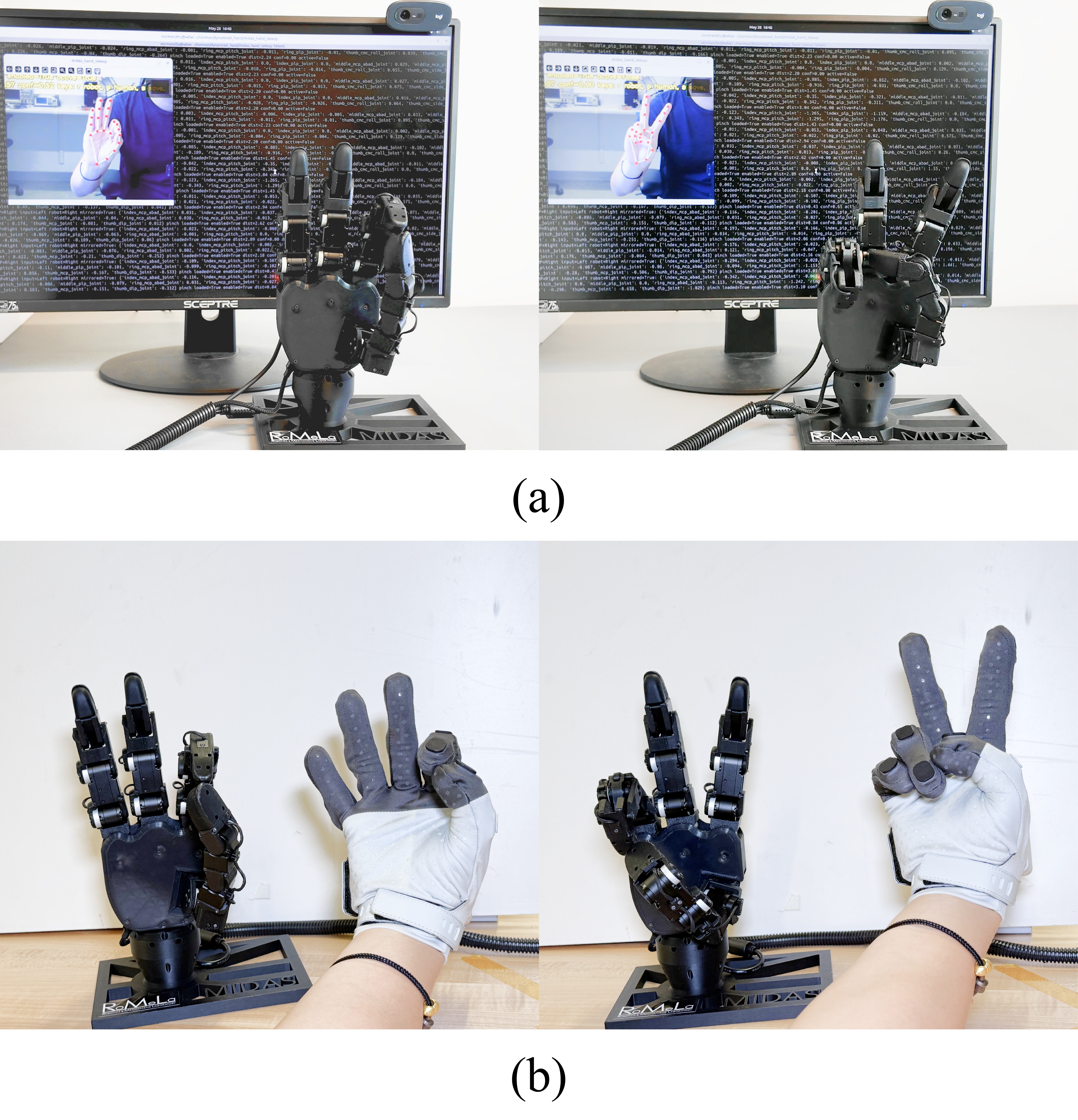}
    \captionsetup{belowskip=-5pt}
    \caption{Retargeting and teleoperation demonstration. Human hand poses are mapped to \ACRO joint commands in real time through (a) vision-based hand tracking and (b) the MANUS glove.}
    \label{fig:teleop}
\end{figure}

The software stack spans the pipeline from human input to data collection. Human finger motion is captured either through vision-based hand tracking or a MANUS glove. When paired with the MANUS Haptic Pro, fingertip forces measured by the tactile arrays are mapped back to per-finger haptic feedback, providing the operator with contact cues during demonstration collection. A retargeting layer performs fingertip inverse kinematics with pinch calibration to convert the tracked motion into robot joint commands, built on existing open-source retargeting codebases ~\cite{qin2023anyteleop} and tuned for the \ACRO kinematics. % TODO: cite the specific retargeting codebase used
As shown in \cref{fig:teleop}, both input modalities enable \ACRO to reproduce different human hand configurations in real time. The underlying hardware API provides actuator joint control together with automatic joint and tactile-sensor calibration. The auto-homing routine re-establishes consistent joint references before use, which is particularly important for a 3D-printed hand, where assembly variation, actuator replacement, and gradual wear of printed components introduce small offsets over time.
 
For simulation, we provide URDF files, MJCF models with simplified contact, and setup files for simulation-based learning environments, supporting visualization, controller development, and sim-to-real policy transfer. All layers lead into a data collection interface that streams synchronized joint and tactile states, enabling demonstration collection for learning-based manipulation with minimal additional setup.
 
\section{Experiments and Results}
\label{sec:Results}

\subsection{Backdrivability}
We quantify backdrivability as the \emph{single-joint backdrive torque}: the joint torque required to initiate motion when a joint is pushed externally. All other joints of the hand are locked so that only the joint under test rotates; a slowly increasing force is applied with a digital force gauge, and the force at motion onset is converted to joint torque through the measured moment arm. Actuators are torque-disabled, so the values reflect passive mechanical backdrivability rather than active compliance. The identical protocol was applied in-house to the Sharpa Wave and Wuji Hand.

\begin{table}[t]
\centering
\caption{Single-joint backdrive torque ($\mathrm{N}\cdot\mathrm{m}$; lower is more backdrivable). Sharpa Wave and \ACRO were measured in-house with the same force-gauge protocol; Wuji Hand joints could not be backdriven.}
\label{tab:backdrive}
\begin{tabular}{llccc}
\toprule
 & Joint & Sharpa Wave & Wuji Hand & \ACRO \\
\midrule 
\multirow{3}{*}{Finger}
 & MCP & 0.35 & $\times$ & \textbf{0.02} \\
 & PIP & 0.11 & $\times$ & \textbf{0.02} \\
 & DIP & 0.03 & $\times$ & passive \\
\midrule
\multirow{3}{*}{Thumb}
 & CMC & 0.62 & $\times$ & \textbf{0.02} \\
 & MP  & 0.32 & $\times$ & \textbf{0.02} \\
 & IP  & 0.07 & $\times$ & \textbf{0.02} \\
\bottomrule
\end{tabular}
\par\vspace{3pt}
{\footnotesize $\times$: joint could not be backdriven.}
\end{table}

Because \ACRO uses identical actuators at every joint, its backdrive torque is essentially uniform at ${\approx}0.02~\mathrm{N}\cdot\mathrm{m}$ (Table~\ref{tab:backdrive}); the finger DIP is passive and inherits the PIP compliance. The Sharpa Wave requires $0.03$--$0.62~\mathrm{N}\cdot\mathrm{m}$, highest at the proximal joints, making \ACRO roughly $3.5$--$30\times$ more backdrivable despite both using directly-driven actuation, while the Wuji Hand joints are non-backdrivable. These results confirm the low joint impedance targeted by our design, enabling compliant, contact-rich interaction; the backdrivable joints also yield under sudden external loads rather than resisting them, improving impact resistance and overall robustness.

\subsection{Thumb Opposition Workspace}

\begin{figure*}
    \centering
    \includegraphics[width=\linewidth]{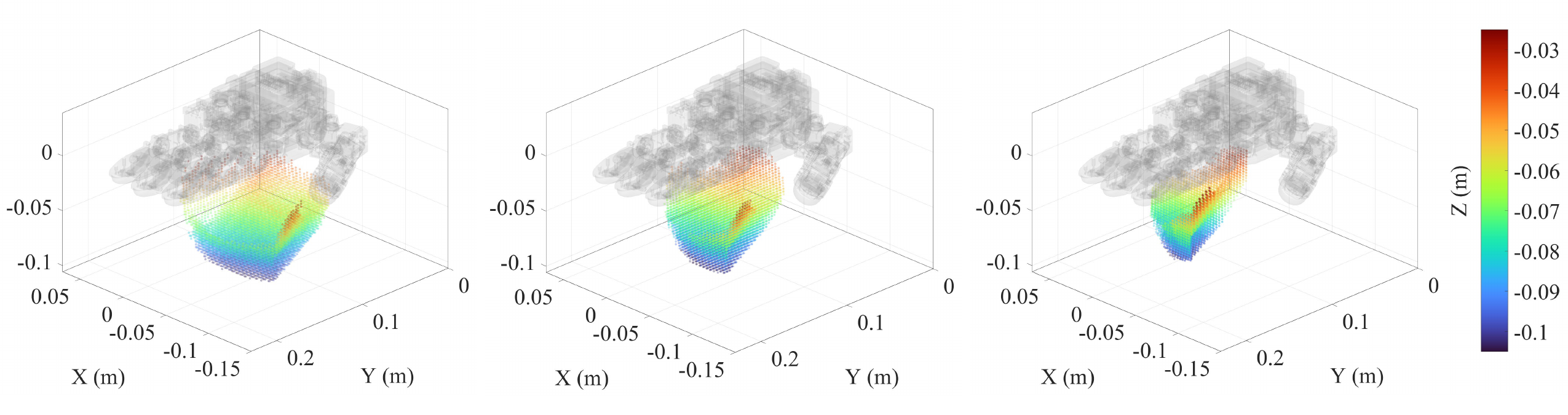}
    \caption{Thumb and finger contact-center workspace intersection for \ACRO. The colored points indicate the overlapping workspace where the thumb and each opposing finger can reach a common contact location.}
    \label{fig:workspace}
\end{figure*}

To evaluate thumb opposability, we computed the intersection between the thumb contact-center workspace and that of each opposing finger, normalized by the corresponding finger workspace: the portion of each finger's reachable workspace that the thumb can oppose. As shown in \cref{fig:workspace} (a), the normalized overlap decreases from the index finger (49.52\%) to the middle (40.03\%) and ring fingers (24.54\%). The measured thumb opposition workspace follows a human-like radial-to-ulnar distribution. Kuo et al.~\cite{KUO2009829} reported that, in normal human hands, the normalized functional workspace for thumb-finger precision manipulation decreases from the index to the middle and ring fingers (33.65\%, 27.10\%, and 23.50\%, respectively). The same decreasing trend observed in our robotic hand suggests that its thumb opposition workspace is similar to that of the human hand in its cross-finger distribution.
 
% Applying the same metric to the LEAP Hand, \cref{fig:workspace} (b) yields normalized overlaps of 39.03\%, 40.23\%, and 28.47\% for the index, middle, and ring fingers. For LEAP Hand, we additionally removed configurations where the fingers intersected the palm. This correction was necessary as the original URDF permits joint configurations where the finger links penetrate the palm geometry, which would otherwise introduce physically infeasible thumb--finger opposition poses. No analogous filtering was required for \ACRO, since its URDF and joint limits limits the fingers to enter the palm region during the sampled motions. While the LEAP Hand also provides substantial thumb-finger opposition, its overlap peaks at the middle finger rather than the index, whereas \ACRO preserves the human radial-to-ulnar organization of thumb-finger opposition for common grasp patterns.
\begin{figure}[htbp!]
    \centering
    \includegraphics[width=0.85\linewidth]{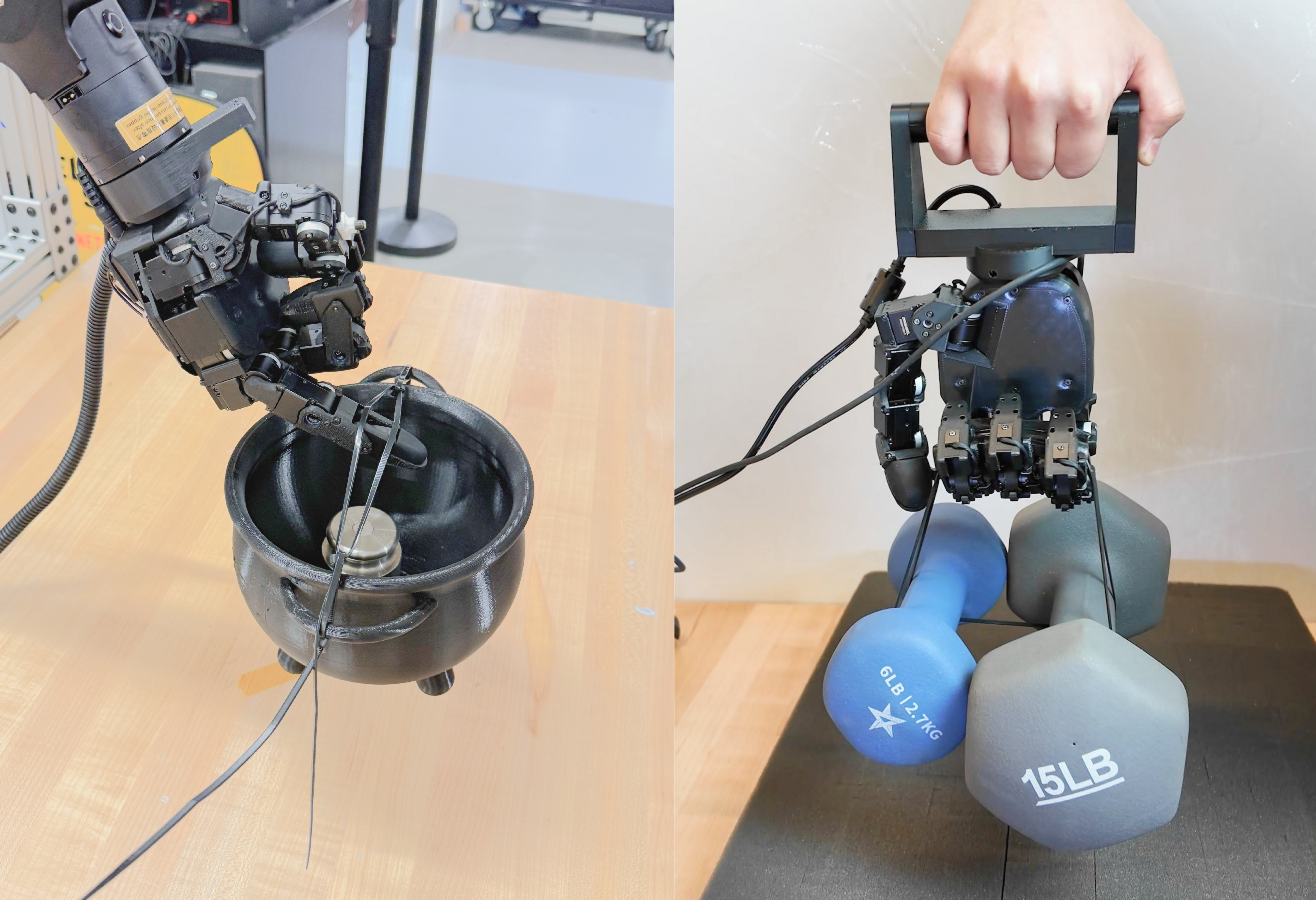}
    \captionsetup{belowskip=-10pt}
    \caption{Strength tests of \ACRO. (a) Fingertip load test. (b) Whole-hand payload test.}
    \label{fig:strength}
\end{figure}

\subsection{Strength Tests}
To evaluate the structural strength and load capacity of \ACRO, we performed two representative loading tests. In the fingertip load test shown in \cref{fig:strength} (a), the index finger was used to lift a suspended load of up to 1.2\,kg, demonstrating the strength of the fingertip and transmission under concentrated loading. The actuator overheated shortly after lifting this load, indicating that sustained high-load operation is limited primarily by actuator thermal limits rather than immediate mechanical failure. In the whole-hand payload test shown in \cref{fig:strength} (b), \ACRO was able to grasp and lift a total load of around 9.5\,kg. The result indicates that the mechanical payload capacity of the hand exceeds 9.5\,kg under this grasping configuration. More aggressive failure-limit testing will be conducted in the future work to determine the true maximum payload capacity. Together, these tests demonstrate that the hardware can withstand substantial fingertip and whole-hand loads for manipulation.

\subsection{Reliability \& Repeatability}
\begin{figure}[t]
 \centering
 \includegraphics[width=\linewidth]{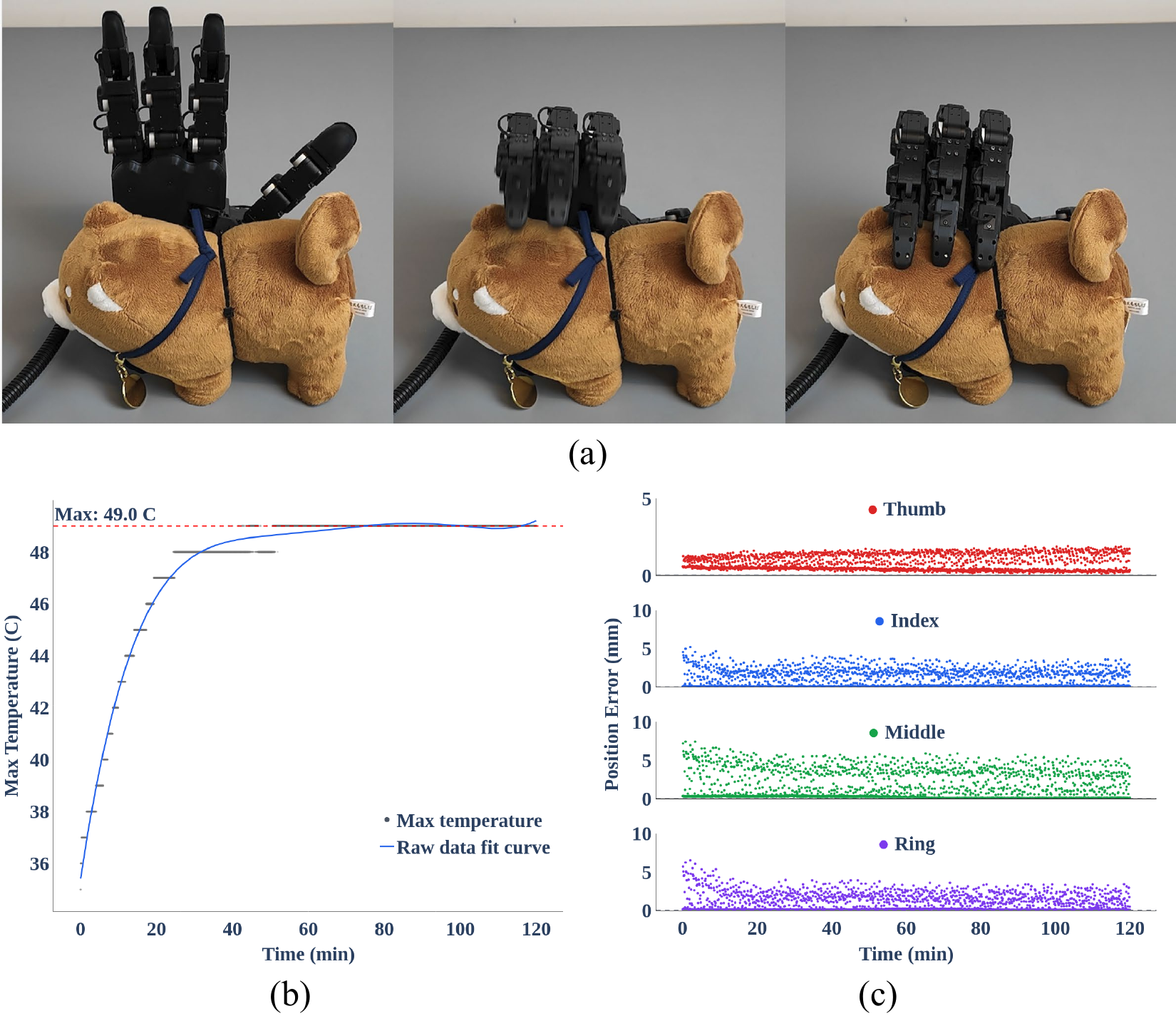}
 \captionsetup{belowskip=-5pt}
 \caption{Reliability test of the \ACRO. (a) Experimental setup for the continuous grasping test. The hand was commanded to repeatedly perform a grasping motion for 2\,h (5,143 cycles). (b) Maximum actuator temperature over time during the test. (c) Endpoint position error over the 2\,h test.}
 \label{fig:reliability}
\end{figure}

With the setup shown in \cref{fig:reliability} (a), we performed a continuous 2-hour grasping test comprising 5,143 grasp cycles to assess the reliability of \ACRO, where actuator temperature and endpoint position error were recorded over time. For each finger, the endpoint position error was computed as the Euclidean distance between the measured fingertip position and its commanded reference position, combining the displacement components in the three directions into a single scalar error value.

As shown in \cref{fig:reliability} (b), the maximum actuator temperature increased rapidly during the initial stage of operation and then gradually converged to a steady value of approximately 49~$^\circ$C, indicating that the system reached thermal equilibrium without continued temperature rise. The endpoint position error of the thumb, index, middle, and ring fingers was also monitored throughout the test, as shown in \cref{fig:reliability} (c). Although small variations were observed, the endpoint errors remained bounded over the full test duration and did not show an obvious increasing trend. These results suggest that \ACRO can maintain stable thermal behavior and consistent finger positioning across more than 5,000 consecutive grasp cycles, supporting its reliability for extended manipulation experiments.

\begin{figure}[t]
    \centering
    \includegraphics[width=\linewidth]{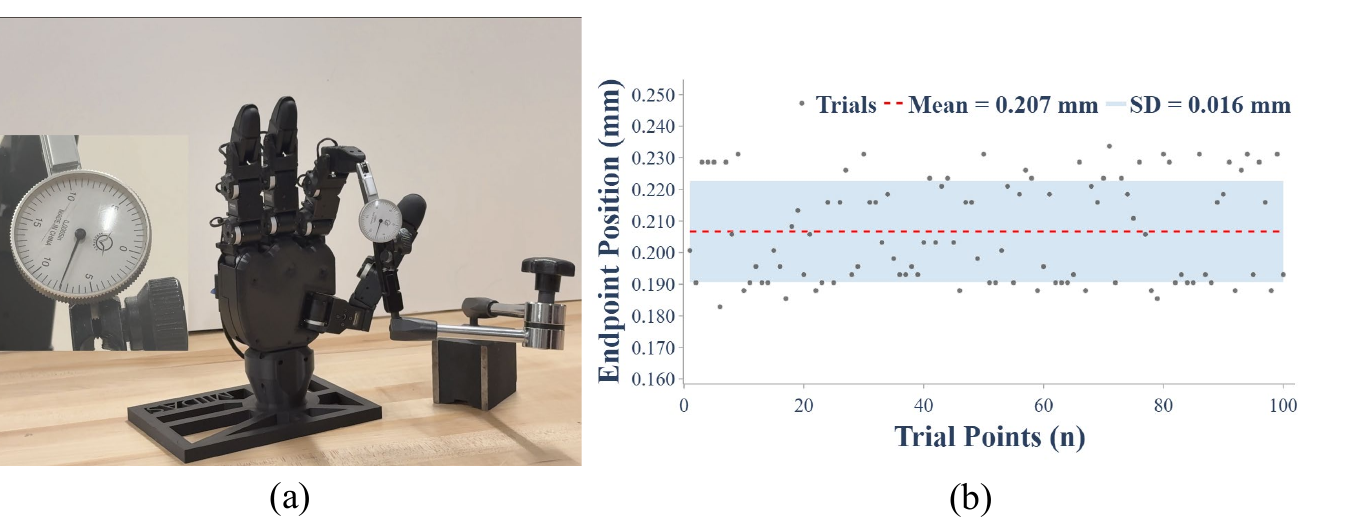}
    \captionsetup{belowskip=-5pt}
    \caption{Repeatability test of the \ACRO. (a) Experiment setup to measure the endpoint position difference for each trial. (b) Experiment result of 100 trials.}
    \label{fig:repeatability}
\end{figure}

To further quantify the finger repeatability, we performed an additional test using a higher-precision measurement setup. The final endpoint position of each closing motion was measured with a mechanical dial indicator over 100 repeated trials (\cref{fig:repeatability} (a)). As shown in \cref{fig:repeatability} (b), the position cluster achieved a repeatability of 0.051\,mm, with average measurement of 0.207\,mm and a standard deviation of 0.016\,mm. This small spread indicates that \ACRO achieves highly consistent repeated closing motions under the tested condition.

\subsection{Dexterity and Grasp Taxonomy}
\begin{figure*}[htbp!]
    \centering
    \includegraphics[width=0.95\linewidth]{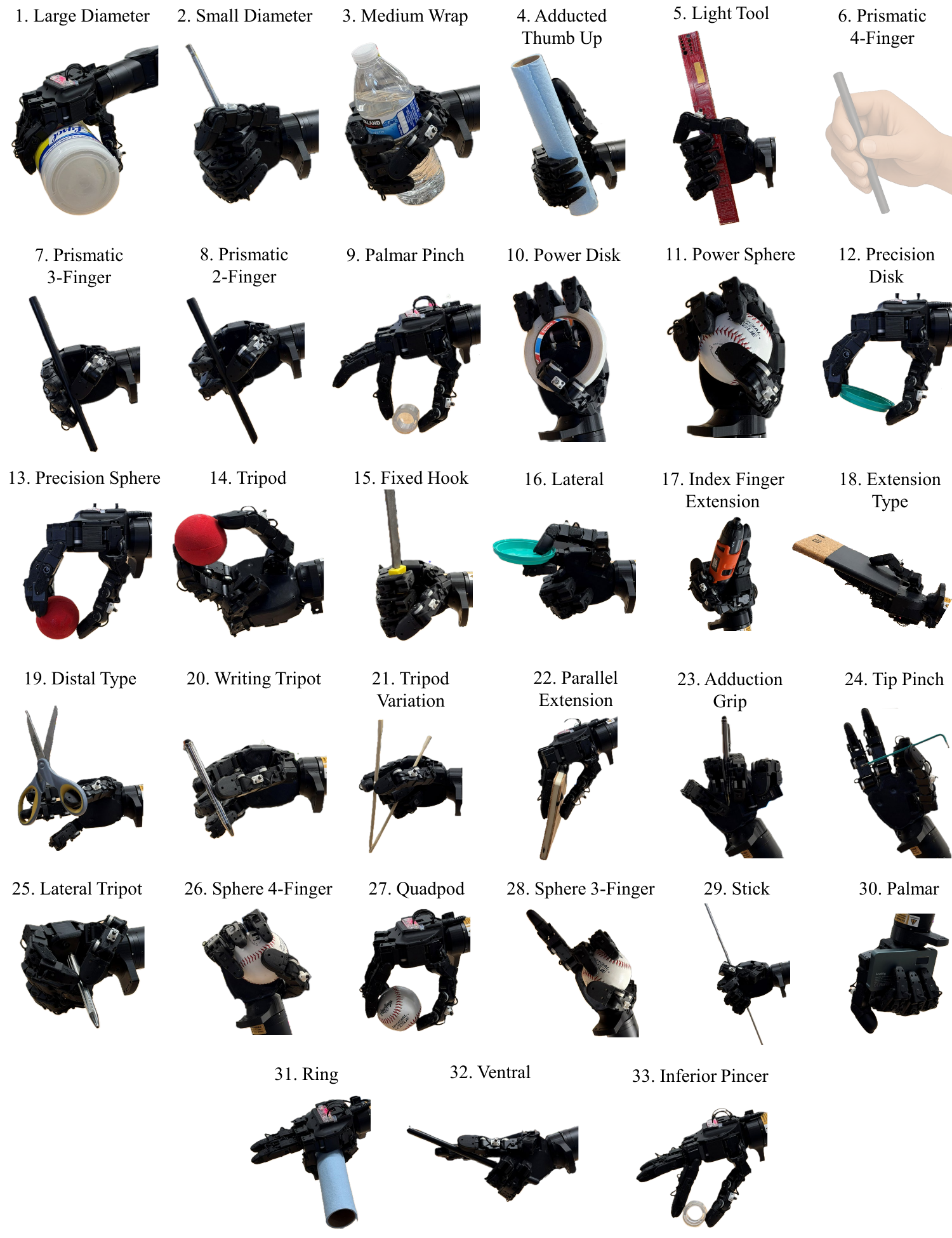}
    \caption{GRASP taxonomy evaluation. \ACRO successfully reproduces 32 of the 33 grasp types defined in the GRASP taxonomy. The only unsuccessful grasp requires the little finger, which is absent in the current four-finger design.}
    \label{fig:taxonomy}
\end{figure*}

To assess the dexterity of \ACRO, we tested the hand on the 33 grasp types defined in the GRASP taxonomy~\cite{feix2016grasp}. As shown in \cref{fig:taxonomy}, \ACRO successfully reproduced 32 of the 33 grasp types, covering a broad range of power, precision, intermediate, and tool-related grasps. These results demonstrate that the hand provides sufficient kinematic dexterity and thumb opposition to support diverse grasping interactions with everyday objects. The only unsuccessful grasp type required the little finger, which is absent in the current four-finger \ACRO design. Overall, the taxonomy evaluation shows that \ACRO covers most common human grasp categories despite a reduced finger.

\section{Discussion}
\label{sec:Discussion}
The results position \ACRO as a practical hardware platform for contact-rich manipulation and clarify its design trade-offs. The uniformly low backdrive torque (${\approx}0.02~\mathrm{N}\cdot\mathrm{m}$, below even the directly-driven commercial hands we measured) enables compliant interaction, and joints that yield under sudden loads add robustness across the strength and reliability tests. The passive DIP coupling trades independent distal control and part of the human sagittal workspace (80.1\% coverage) for a lower actuator count and cost, yet the 32 of 33 taxonomy grasps and the human-like opposition trend indicate that the dexterity needed for common grasps is preserved. Finally, payload is bounded by actuator thermal limits rather than structural strength: the 3D-printed structure sustained a 9.5\,kg whole-hand load while a 1.2\,kg fingertip load overheated the actuator, and continuous operation stabilized near 49~$^\circ$C, providing practical duty-cycle guidance for sustained experiments.

Several limitations remain. The four-finger design simplifies the hand and lowers cost, but the absence of a little finger limits grasps that rely on ulnar-side support, as reflected in the taxonomy test, and may reduce stability in some in-hand manipulation motions. The current evaluation characterizes hardware-level properties rather than task-level autonomous manipulation, so the results should be read as evidence of hardware readiness rather than manipulation performance. In addition, tactile sensing is presently demonstrated as a sensing modality and has not yet been incorporated into closed-loop control. Future work will quantitatively characterize the integrated tactile system, incorporate tactile feedback into control, extend the software ecosystem with learning interfaces, and evaluate manipulation policies trained with and without tactile input on the released platform.

\section{Conclusion}
\label{sec:Conclusion}
We presented \ACRO, a low-cost, open-source, human-scale dexterous hand that combines directly-driven backdrivable actuation, dense three-axis tactile sensing, and modular 3D-printed construction in a single reproducible platform. Experiments demonstrated a uniformly low backdrive torque of ${\approx}0.02~\mathrm{N}\cdot\mathrm{m}$, roughly $3.5$--$30\times$ lower than a comparable directly-driven commercial hand; coverage of 32 of the 33 GRASP taxonomy types; a human-like radial-to-ulnar thumb opposition trend; whole-hand payloads exceeding 9.5\,kg; closing repeatability with a standard deviation of 0.016\,mm; and stable operation over more than 5,000 consecutive grasp cycles. Together with the released hardware designs, control and tactile APIs, simulation models, and teleoperation pipeline, these results establish \ACRO as a practical foundation for tactile feedback control, scalable demonstration collection, and learning-based dexterous manipulation.

% {
% \bibliographystyle{IEEEtran}
% \bibliography{references}
% }
\renewcommand*{\bibfont}{\footnotesize}
\printbibliography

\end{document}